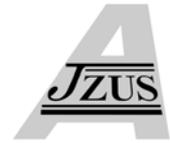

# Wavelet-based deconvolution of ultrasonic signals in nondestructive evaluation[*]

HERRERA Roberto Henry[†1], OROZCO Rubén[2], RODRIGUEZ Manuel[3]

(*[1]Department of Informatics, University of Cienfuegos, Cienfuegos 59430, Cuba*)
(*[2]CEETI, Central University of Las Villas, Santa Clara 54830, Cuba*)
(*[3]Department of Physics, Central University of Las Villas, Santa Clara 54830, Cuba*)
[†]E-mail: henry@finf.ucf.edu.cu
Received Mar. 1, 2006; revision accepted July 17, 2006

**Abstract:** In this paper, the inverse problem of reconstructing reflectivity function of a medium is examined within a blind deconvolution framework. The ultrasound pulse is estimated using higher-order statistics, and Wiener filter is used to obtain the ultrasonic reflectivity function through wavelet-based models. A new approach to the parameter estimation of the inverse filtering step is proposed in the nondestructive evaluation field, which is based on the theory of Fourier-Wavelet regularized deconvolution (ForWaRD). This new approach can be viewed as a solution to the open problem of adaptation of the ForWaRD framework to perform the convolution kernel estimation and deconvolution interdependently. The results indicate stable solutions of the estimated pulse and an improvement in the radio-frequency (RF) signal taking into account its signal-to-noise ratio (SNR) and axial resolution. Simulations and experiments showed that the proposed approach can provide robust and optimal estimates of the reflectivity function.

**doi:**10.1631/jzus.2006.A1748     **Document code:** A     **CLC number:** TB55

## INTRODUCTION

Deconvolution of ultrasonic signals is defined as the solution of the inverse problem of convolving an input signal, known as the system function $h(n)$, with a medium reflectivity function $x(n)$ and can be represented by

$$y(n)=h(n)*x(n)+\eta(n), \qquad (1)$$

where $y(n)$ is the measured signal, $*$ denotes the convolution operation and $\eta(n)$ is the additive noise.

Recovering $x(n)$ from the observation $y(n)$ leads to improving the appearance and the axial resolution of the RF-signals by removing the dependent effects of the measuring system (Wan et al., 2003). The signal $y(n)$ corresponds to RF lines of 2D acoustic image or 1D signal, where the problem is addressed by taking the desired signal $x(n)$ as the input of a linear time invariant (LTI) system with impulse response $h(n)$. The output of the LTI system is blurred by Gaussian white noise $\eta(n)$ of variance $\sigma^2$. In the frequency domain Eq.(1) can be expressed as

$$Y(f)=H(f)X(f)+N(f), \qquad (2)$$

where $Y(f)$, $H(f)$ and $N(f)$ are the Fourier transformations of $y(n)$, $h(n)$ and $\eta(n)$, respectively.

If the system frequency response $H(f)$ has no zeros, an estimation of $x(n)$ can be obtained by using a simple inverse filter (Neelamani et al., 2004). However, where $H(f)$ takes values close to zero, the noise is highly amplified, leading to incorrect estimates. In such case it is necessary to include a regularization

[*] Project (No. PRC 03-41/2003) supported by the Ministry of Construction of Cuba

parameter in the inverse filter, which reduces the variance of the estimated signal. The best-known example of regularized filter for stationary signals is the Wiener filter (Adam and Michailovich, 2002).

When the signals under analysis show non-stationary properties, such as abrupt changes, the Wiener filter based on the Fourier transform does not give satisfactory results in the estimation. This is conditioned by the characteristics of the Fourier basis of complex exponentials (Wan et al., 2003). It is known that wavelets orthonormal basis achieves a better matching with the transmitted pulse and leads to a better localization in time and frequency domain (Neelamani et al., 2004). One of the advantages of wavelets is that the signals can be represented with some few coefficients different from zero, which corresponds with the sparse characteristics of the ultrasonic signals, where the trace is only composed by values different from zero in cases of abrupt changes of acoustic impedance.

Neelamani et al.(2004) proposed a wavelet-based regularized deconvolution technique (ForWaRD) to solve the inverse filtering process, which will be used in this paper for the deconvolution of ultrasonic signals; furthermore, we adapt the ForWaRD, by adding the pulse estimation as a prior step of the ForWaRD process.

The initial problem in deconvolution is the existing or absence of prior knowledge of the system impulse response $h(n)$. Oppenheim and Schafer (1989) defined the case of estimating $x(n)$ from $h(n)$ as the well-known homomorphic deconvolution using the real cepstrum for minimum phase signals, or the complex cepstrum for the most general case. Taxt (1997) compared seven cepstrum-based methods for blind deconvolution in the estimation of the reflectivity function in biological media. Michailovich and Adam (2003) proposed wavelet-based projection methods (WBPM) and wavelet-based denoising methods, and achieved significant results in the estimation. However, in all these methods some previous knowledge of the system is required, or the assumption of minimal-phase pulse must be considered. This is something that depends on the construction of the housing of the piezo-electric and of the impedance matching between the transmitter and the crystal.

We have selected the method of higher order spectral analysis (HOSA) because of its immunity to noise, and to avoid the assumption of minimum phase for the transducer's electromechanical impulse response.

The process is divided into two stages. The first one is the estimation of the ultrasound pulse from the bicepstrum. Once an estimation of this function is obtained, it is used to cancel out the blurring effect of the pulse from the observation data by the selected deconvolution procedure.

ULTRASOUND PULSE ESTIMATION

The system function described in Eq.(1) as the transducer impulse response $h(n)$ is a deterministic and causal FIR filter, $x(n)$ representing the medium response function assumed by the authors initially, without loss in generality, stationary, zero mean and non-Gaussian distribution. This last property guarantees that its third-order cumulant exists, with $\eta(n)$ representing the zero mean Gaussian noise which is uncorrelated with $x(n)$. The third-order cumulant of the zero mean signal $y(n)$ is represented by Abeyratne et al.(1995)

$$c_3^y(m_1,m_2) = \frac{1}{M}\sum_{k=0}^{M-1} y(k)y(k+m_1)y(k+m_2), \quad |m_1| \leq L, \ |m_2| \leq L, \quad (3)$$

where $L$ is the correlation lag.

From Eq.(3), the bicepstrum can be computed by using the algorithm reported in (Pan and Nikias, 1988)

$$b_y(m_1,m_2) = F_{2D}^{-1}\left\{\frac{F_{2D}[m_1 c_3^y(m_1,m_2)]}{F_{2D}[c_3^y(m_1,m_2)]}\right\}, \quad (4)$$

where $F_{2D}$ and $F_{2D}^{-1}$ are the direct and inverse bidimensional Fourier transform respectively.

The cepstrum $\hat{h}(n)$ of $h(n)$ is obtained by evaluating the bicepstrum along the diagonal $m_1=m_2$ for all $n \neq 0$ (Abeyratne et al., 1995)

$$\hat{h}(n) = b_y(-n,n), \quad \forall n \neq 0. \quad (5)$$

Then, from Eq.(5) we may estimate $h(n)$ as (Oppenheim and Schafer, 1989)

$$h(n) = F^{-1}\left\{\exp\left[F(\hat{h}(n))\right]\right\}. \tag{6}$$

The bispectrum of the white Gaussian noise is zero, which allows estimation of $h(n)$ without taking into account the contribution of $\eta(n)$ in Eq.(1).

## WIENER INVERSION

Once the pulse is estimated to perform deconvolution the Wiener filter is used. The common accepted expression to obtain the estimation from Eq.(2) is presented in (Honarvar et al., 2004) as follows:

$$X_{est}(f) = \frac{Y(f)H^*(f)}{|H(f)|^2 + Q^2}, \tag{7}$$

where $X_{est}(f)$ is the Fourier transform of the estimated reflectivity function, and $Q^2$ is the regularization parameter, which is properly selected to control the noise content and to avoid the indetermination of Eq.(7).

In (Honarvar et al., 2004), $Q^2$ was called noise desensitizing factor and established as: $Q^2 = \max(|H(f)|^2)/100$, which uses a fixed value as regularization parameter. Instead of this approach, we used a median absolute deviation (MAD) estimator, following the method proposed by Donoho (1995), which estimates the noise variance on the finest scale wavelet coefficients of the observation $y(n)$.

## ForWaRD IMPLEMENTATION

Reflectivity function estimation by means of Eq.(7) implies a Fourier shrinkage. Neelamani et al. (2004) proposed another step, so that the signal with less Fourier representation may be treated by shrinking its wavelet coefficients, leading to better representation of both, smooth signals and highly spiky signals, as it is expected from the ultrasonic reflectivity function.

**ForWaRD algorithm**

Given the observation $y(n)$ and the ultrasound pulse $h(n)$, and setting them to pairs of wavelets functions $(\phi_1,\psi_1)$ and $(\phi_2,\psi_2)$, the first one is used for the denoising stage and Fourier shrinkage. Then, the second one is used for wavelet shrinkage and inverse wavelet transformation. The ForWaRD algorithm can be summarized as:

(1) Noise variance estimation $\sigma_\eta$:

Discrete wavelet transform (DWT) of the observation $y$, to obtain the detail coefficients $d_L$, on the finest decomposition level $L$. The noise standard deviation $\sigma_\eta$ is obtained by using an MAD estimator, as was proposed by Donoho (1995)

$$\sigma_\eta = Median(|d_L|)/0.6745. \tag{8}$$

(2) The regularization parameter $\tau$ for Tikhonov shrinkage was determined by Neelamani et al.(2004), to achieve the best results in the mean square error (MSE) sense, in the range $[0.01 \sim 10] N\sigma_\eta^2/|y-\mu_y|^2$, where $N$ is the length of $y$, and $\mu_y$ its mean.

(3) The first estimation of the reflectivity function is as follows:

$$X_{\lambda 1}(f) = \frac{Y(f)}{H(f)}\left[\frac{|H(f)|^2}{|H(f)|^2 + \tau}\right], \tag{9}$$

where the term in brackets is the shrinking function in the Fourier domain.

(4) The fourth step involves only an inverse Fourier transformation (IFT) to obtain the first estimation of the reflectivity function.

(5) The estimated reflectivity function is decomposed in the wavelet domain by using the pair $(\phi_1,\psi_1)$ in the DWT for denoising purpose. The wavelet $\psi_1$ must be selected between those which best represent the ultrasound pulse $h$.

(6) This step is similar to the previous one, but uses the pair $(\phi_2,\psi_2)$, with a smooth wavelet, i.e. only few vanishing moments. Finally, we obtained $a_{2,j}$ and $d_{2,j}$, the approximation and detail coefficients respectively.

(7) A level dependent threshold is applied over the detail coefficients $d_{1,j}$ (Donoho, 1995)

$$T_j = \sigma_j \sqrt{2\lg N}, \tag{10}$$

where $\sigma_j$ is the noise standard deviation at each decomposition level and $N$ is the number of samples. The hard thresholding is selected to obtain $d_{th,j}$

(Neelamani *et al.*, 2004).

(8) The shrinkage in the wavelet domain $\lambda_j^w$ is expressed as

$$\lambda_j^w = \frac{|d_{th,j}|^2}{|d_{th,j}|^2 + \sigma_j^2}. \quad (11)$$

(9) Eq.(11) is used to shrink the detail coefficients obtained from Step (8). This is an extension of Wiener filter to the wavelet domain, which is obtained by

$$d_{sh,j} = d_{2,j} \lambda_j^w. \quad (12)$$

(10) The final step is the inverse discrete wavelet transformation (IDWT) by using the approximations $a_{2,j}$, the detail coefficients obtained from Eq.(12) and the pair $(\phi_2, \psi_2)$ to obtain the reflectivity function $x_F(n)$.

The pseudo-code description is as follows:
(1) DWT $\{y, (\phi_1, \psi_1)\} \rightarrow$ estimate $\sigma_\eta$ using MAD;
(2) Obtain $\tau$ using $\sigma_\eta$ and $y$;
(3) $X_{est} = Y/H$ and shrink $X_{est}$ using $\tau \rightarrow X_{\lambda 1}$;
(4) IFT $\{X_{\lambda 1}\} \rightarrow x_{\lambda 1}$;
(5) DWT $\{x_{\lambda 1}, (\phi_1, \psi_1)\} \rightarrow a_{1,j}, d_{1,j}$;
(6) DWT $\{x_{\lambda 1}, (\phi_2, \psi_2)\} \rightarrow a_{2,j}, d_{2,j}$;
(7) Apply hard thresholding to $d_{1,j} \rightarrow d_{th,j}$;
(8) Calculate $\lambda_j^w$ using $d_{th,j}$ and $\sigma_j$;
(9) Shrink $d_{sh,j} = d_{2,j} \lambda_j^w$;
(10) IDWT $\{a_{2,j}, d_{sh}, (\phi_2, \psi_2)\} \rightarrow x_F(n)$.

## COMPUTER SIMULATIONS

### Pulse estimation

For comparing purposes in this section we used the same simulated ultrasound pulse as in (Adam and Michailovich, 2002), which is a damped sine function generated according to

$$h(k) = k \sin(2\pi \cdot 0.07k) \exp\left[-0.005\left(k - \frac{P}{2}\right)^2\right], \quad (13)$$

where $k=1, ..., P$, with $P$ being the pulse length. Instead of a Gaussian reflectivity function used in the cited study, we used a Bernoulli-Gaussian distribution to represent $x(n)$ as was defined in (Kaaresen and Bolviken, 1999) a logical computation with a Gaussian white noise (GWN), $[GWN < \rho_r] \cdot GWN$, where the probability $\rho_r$ governs the sparsity of the distribution. We have called $\rho_r$ as density of reflectors.

The first simulation is based on a reflectivity function (1024 samples) with $\rho_r = 0.03$.

The convolution process is depicted in Fig.1, where the generated observation $y$ is contaminated with GWN distribution $N(0, \sigma_\eta^2)$, while $\sigma_\eta^2$ is selected to have a signal to noise ratio $SNR=14$ dB, and its calculation is given by $SNR=20\lg(\|y\|/\|y-y_\eta\|)$, where $y_\eta$ is the noisy signal.

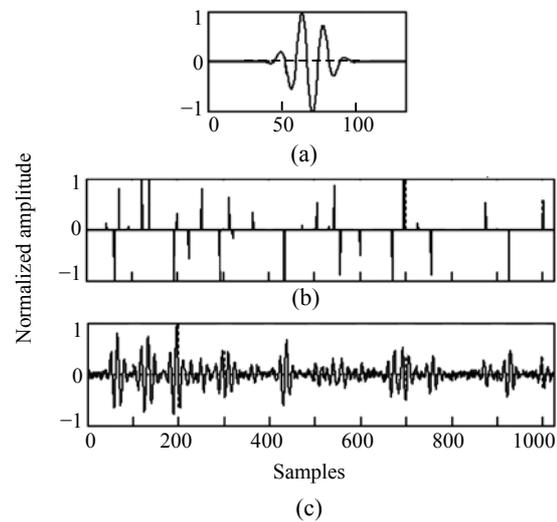

**Fig.1 Convolution process with *SNR*=14 dB. (a) Ultrasound pulse *h*; (b) Reflectivity function *x*, with $\rho_r$=0.03; (c) Observation *y***

The pulse estimation using the described model was achieved over 50 randomly generated reflectivity functions with the same $\rho_r$, with the results being shown in Fig.2, where all the estimated pulses in the upper plot are superimposed; and are shown in the lower plot, the mean in solid line, and ±2 standard deviation (SD) in dotted line.

The same procedure was followed using an $SNR=10$ dB, and 7 dB to test the performance of the algorithm in different situations as it is found in non-destructive evaluation (NDE). The results are shown in Table 1, where the mean square error (MSE) of the estimation is used to quantify the quality of the estimation.

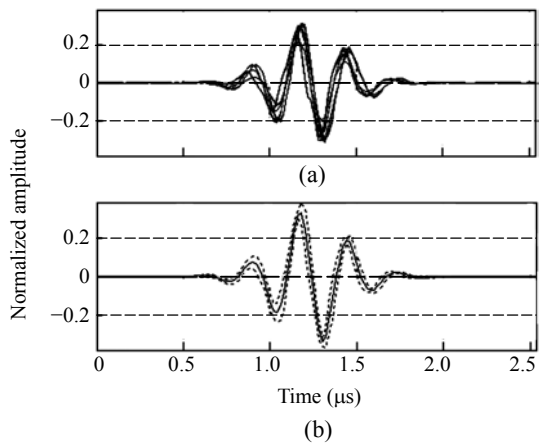

**Fig.2 Pulses estimated using HOSA. (a) Estimated pulses; (b) Mean±2 SD**

**Table 1 MSE results**

| SNR (dB) | HOSA [MSE±2 SD] | W+D[*] [MSE±2 SD] |
|---|---|---|
| 14 | 0.043±0.031 | 0.034±0.030 |
| 10 | 0.059±0.038 | 0.045±0.034 |
| 7 | 0.082±0.058 | 0.086±0.051 |

[*] W+D results given in (Michailovich and Adam, 2003)

The W+D method in Table 1 is the wavelet-based denoising method proposed by Michailovich and Adam (2003). From these results we concluded that W+D outperforms the HOSA based method for high SNR values, while the HOSA shows a small improvement in the MSE at *SNR*=7 dB. In spite of the fact that the statistical difference is not significant, the HOSA model does not have the restriction of minimum phase of the other methods. Both cases in Table 1 outperform the cepstrum and wavelet-based projections methods reported in (Michailovich and Adam, 2003).

**Reflectivity function estimation**

To demonstrate the performance of the ForWaRD algorithm we use a simulated reflectivity function with a low density of reflectors $\rho_r$=0.01. After the convolution a GWN was added to the resultant signal so as to obtain an *SNR* of 14 dB (shown in Fig.3).

The deconvolution algorithm must be able to distinguish between near reflectors, as shown in the reflectivity function in Fig.3. See samples from 450 to 500; 750 to 800, where the pulses are overlapped, and from 950 to 1000. Some of these reflections have an inverted phase which must be resolved by the ForWaRD algorithm. Initially the pulse was estimated by using HOSA. The conventional Wiener filter was regularized with $Q^2$ followed by autoregressive spectral extrapolation (ASE) as in (Honarvar et al., 2004) with a Burg model of order 20.

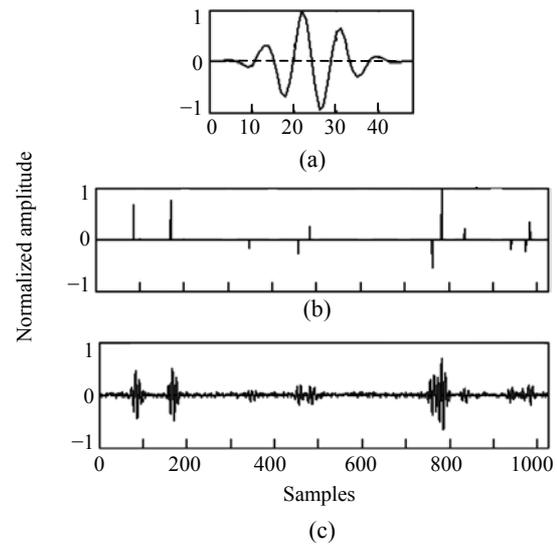

**Fig.3 Simulated observation with $\rho_r$=0.01 and *SNR*=14 dB. (a) Ultrasound pulse; (b) Reflectivity function *x*, with $\rho_r$=0.01; (c) Observation *y***

The result is shown in Fig.4, where we included the simulated reflectivity function traced to facilitate the visual comparison. In Fig.4b the Wiener filter was applied; the result in Fig.4c was obtained following the model proposed by Honarvar et al.(2004). This result is outperformed by the application of the subsequent step of Wiener filtering in the wavelet domain as it is shown in Fig.4d. We used the Daubechies wavelets DB12 and DB6 as the pairs described to implement the ForWaRD denoising and inversion stages respectively.

The quality of the estimation is measured as the reduction of the time support of the ultrasound pulse, i.e. the axial resolution, and it is evaluated as the autocovariance of the RF signals before and after deconvolution calculated from the envelope. Fig.5 shows the main lobe of the covariance function for the two methods tested in this study; the axial resolution gain is taken as the difference in number of samples at −3 dB drop as in (Adam and Michailovich, 2002). The original RF signal in dotted line has a −3 dB width of 16 samples; the processed signal using the regularized Wiener filter, in dashed line, has 12 samples; while the deconvolved signal using ForWaRD, in solid line, has 7 samples at the same measure. Thus, the axial resolution gain achieved by WienerQ is 1.33, and for the implemented method using ForWaRD is 2.31.

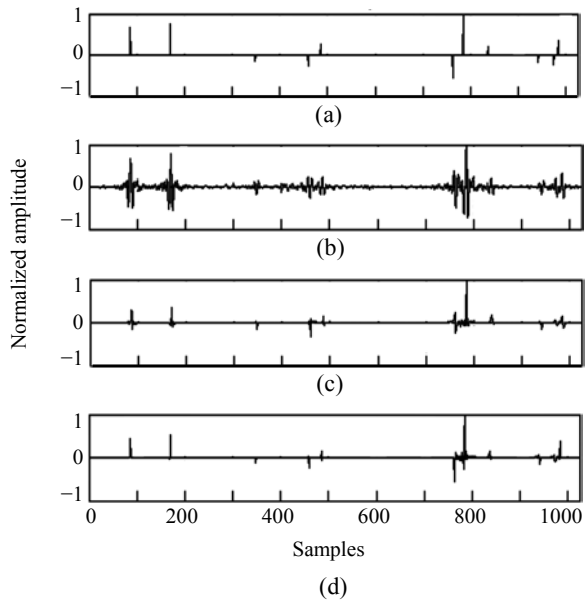

**Fig.4 Deconvolution methods. (a) Simulated reflectivity function; (b) Signal obtained by Wiener filter; (c) Signal obtained by Wiener filter+ASE; (d) Signal obtained by ForWaRD+ASE**

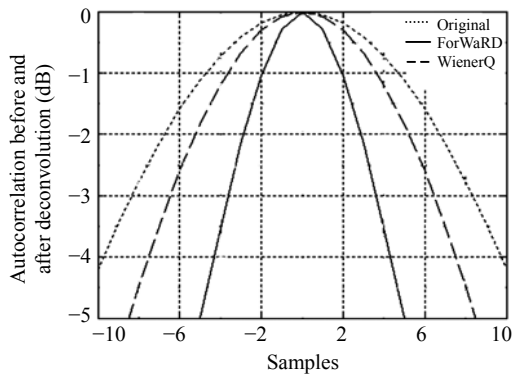

**Fig.5 Autocorrelation of the envelope of original and deconvolved signals**

As in (Neelamani et al., 2004) we evaluated the improvement in SNR (ISNR). In the presented example this value was 0.12 dB and 3.29 dB for WienerQ and ForWaRD methods respectively.

The ForWaRD algorithm was tested with the inclusion of the pulse estimation step. The results in a simulated signal have been successful because the Bernoulli-Gaussian distribution has been carefully calculated with a value of $\rho_r$ which guarantees that it is non-Gaussian or symmetrically distributed. In other cases the application of the bicepstrum concept fails.

The Hinich's criterion (Swami et al., 1998) was used as Gaussianity test providing an upper limit of $\rho_r$=0.2, starting from which the process becomes Gaussian. In real NDE data the reflectivity will be sparce (Kaaresen and Bolviken, 1999) and thus far from Gaussian.

## PROCESSING REAL NDE DATA

**Experimental results for pulse estimation**

All experiments were carried out with a Krautkramer-Branson USN 52R as transmitter. The received signal was acquired with a Tektronix TDS 220 digital oscilloscope connected to a personal computer by an RS232 interface. In this experiment a 16 mm diameter Krautkramer unfocused ultrasonic transducer able to transmit a pulse at 2.25 MHz was used. The international test block (STBA1, see Fig.9) in a cross-section of 25 mm was tested to obtain various back wall reflections, as is depicted in Fig.6 with its power spectrum.

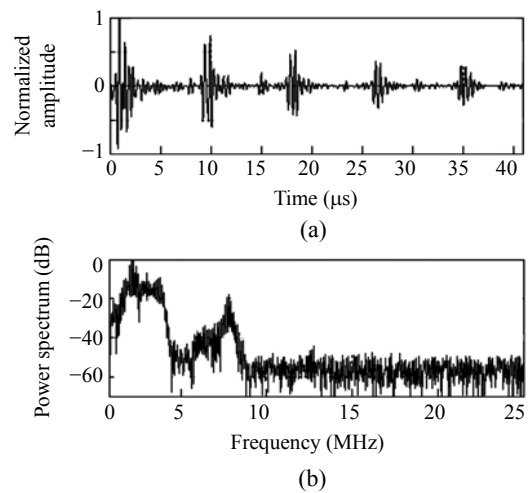

**Fig.6 Signal from STBA1 and power spectrum. (a) Acquired signal; (b) Signal power spectrum**

The sampling frequency was 50 MHz and the estimated velocity in the medium was 5931 m/s. The acquired sequence of 2500 samples length was trun-cated to 2048 samples. For the pulse estimation process this signal was divided into 16 non-overlapping segments with 128 samples for each segment, and the same length was considered for the Fourier transform.

We can clearly see from Fig.7 that the frequency spectrum of estimated pulse has the salient features contained in the second back wall echo measured experimentally. The center frequency of the transducer is 2.25 MHz represented by a peak in the power spectrum plot. The real MSE for our estimated pulse is 0.0682.

**Deconvolution of pulses**

With the estimated pulse the deconvolution procedure is applied to a sequence of multiple reflections from the 25 mm thick section of the STBA1, whose result is shown in Fig.8, where two methods are considered. The WienerQ is related to the Wiener filtering using the parameter $Q^2$ as has been used in the simulations, and the ForWaRD is our implementation of the algorithm proposed by Nelamani *et al.*(2004).

Location and phase of all the three echoes (Fig.8) were detected by both methods. Although, the estimated reflectivity function using the approach proposed in this paper shows better resolved peaks quantified by axial resolution gain of 2.18 and an improvement in SNR (ISNR) of 5.6 dB. Whereas, these values for the WienerQ were 1.88 dB and 2.4 dB respectively.

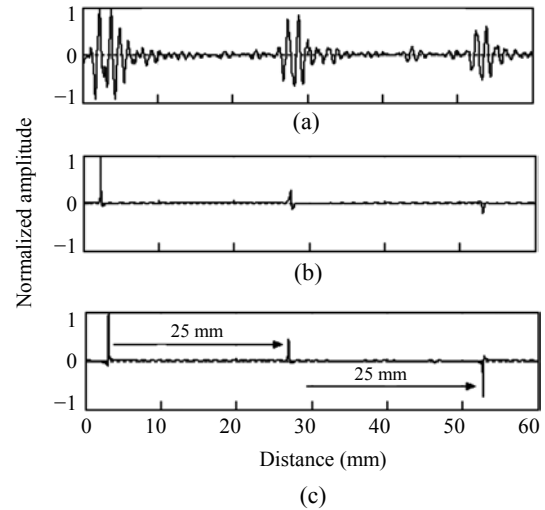

**Fig.8 Deconvolution of three back wall echoes. (a) Three echoes from the backwall; (b) Estimated reflectivity using WienerQ; (c) Estimated reflectivity using ForWaRD**

**Resolving closely positioned reflectors**

Using the same experimental conditions in the STBA1 the depth of the notch was tested to seek near reflections to be separated as is depicted in Fig.9.

The signal acquired in the position shown in Fig.9 presupposes a difficult test for the algorithm, two reflections are overlapped (*H*2 and *H*3) while a positive phase is expected for all reflections. Again, the same pairs of wavelets were used and the ForWaRD deconvolution procedure was followed by ASE. The estimated pulse by HOSA was used as the wavelet for deconvolution.

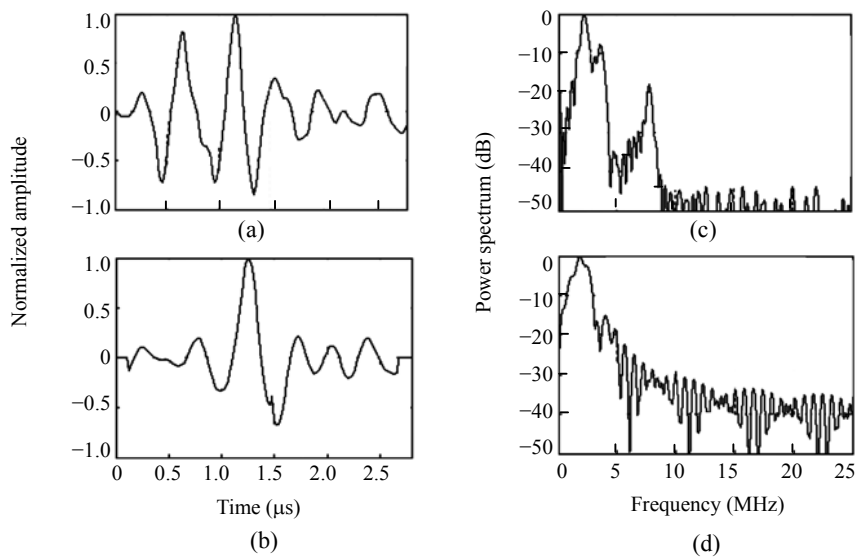

**Fig.7 Estimated pulse and the power spectrum. (a) Second reflection pulse; (b) Estimated pulse; (c) Power spectrum of the second reflection pulse; (d) Power spectrum of the estimated pulse**

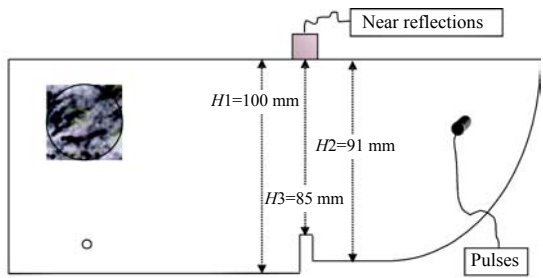

**Fig.9 Diagram of the STBA1 for the experiments**

Fig.10 shows the initial pulse corresponding to the reflection in the front wall; the three sharp peaks located at 85, 91 and 100 mm, with positive phase, are the echoes at depths $H1$, $H2$ and $H3$ in Fig.9. The axial resolution gain for this experiment was 2.33 with an $ISNR$=7.86 dB.

The hypothesis of non-Gaussianity was accepted for all the real NDE signals under test. This was expected as only a limited number of samples had nonzero values.

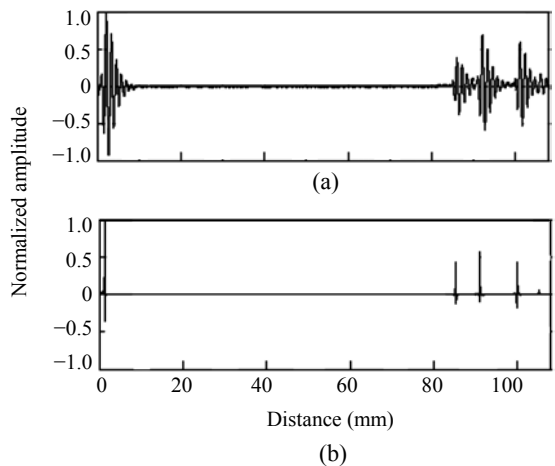

**Fig.10 Estimated reflectivity function from three closely positioned echoes. (a) Signal from three near reflectors; (b) Estimated reflectivity function**

CONCLUSION

In this paper, we introduced a new method for deconvolution of ultrasonic signals in NDE based on the ForWaRD concepts. The main goal of our approach is to obtain the reflectivity function from the blurred observation in a blind deconvolution framework. The goal was successfully achieved, with the resolution gain in simulations and experiments showing the feasibility of the method proposed in this work.

Some problems related to the selection of the appropriate wavelet pairs may be overcome by the use of adapted wavelets (Mesa, 2005) to the estimated ultrasound pulse. The algorithm proved its robustness using the same wavelets in both simulated and real data.

The non-Gaussianity assumption of the RF signal is supported by the success in the pulse estimation stage being tested with Hinich's criterion. Furthermore, the use of HOSA-based models avoids the restriction of minimum phase for the ultrasound pulse as needed in cepstrum-based methods and WBPM. The sparse properties of reflectors were successfully represented using the Bernoulli-Gaussian distribution in the computer generated RF sequences.

**References**

Abeyratne, U.R., Petropulu, A.P., Reid, J.M., 1995. Higher order spectra based deconvolution of ultrasound images. *IEEE Trans. Ultrason. Ferroelect. Freq. Contr.*, **42**(6): 1064-1075. [doi:10.1109/58.476550]

Adam, D., Michailovich, O., 2002. Blind deconvolution of ultrasound sequences using nonparametric local polynomial estimates of the pulse. *IEEE Transactions on Biomedical Engineering*, **49**(2):118-131. [doi:10.1109/10.979351]

Donoho, D.L., 1995. De-noising by soft-thresholding. *IEEE Trans. Inform. Theory*, **41**(3):613-627. [doi:10.1109/18.382009]

Honarvar, F., Sheikhzadeh, H., Moles, M., Sinclair, A.N., 2004. Improving the time-resolution and signal-to-noise ratio of ultrasonic NDE signals. *Ultrasonics*, **41**(9):755-763. [doi:10.1016/j.ultras.2003.09.004]

Kaaresen, K.F., Bolviken, E., 1999. Blind deconvolution of ultrasonic traces accounting for pulse variance. *IEEE Trans. Ultrason. Ferroelect. Freq. Contr.*, **46**(3):564-573. [doi:10.1109/58.764843]

Mesa, H., 2005. Adapted wavelets for pattern detection. *Lecture Notes in Computer Science*, **3773**:933-944.

Michailovich, O., Adam, D., 2003. Robust estimation of ultrasound pulses using outlier-resistant de-noising. *IEEE Trans. on Medical Imaging*, **22**(3):368-381. [doi:10.1109/TMI.2003.809603]

Neelamani, R., Choi, H., Baraniuk, R., 2004. ForWaRD: fourier-wavelet regularized deconvolution for ill-conditioned systems. *IEEE Trans. Ultrason. Ferroelect. Freq. Contr.*, **52**(2):418-432.

Oppenheim, A.V., Schafer, R.W., 1989. Discrete Time Signal Processing. London Prentice-Hall, p.768-825.

Pan, R., Nikias, C., 1988. The complex cepstrum of higher order cumulants and nonminimum phase system identification. *IEEE Trans. ASSP*, **36**(2):186-205.

Swami, A., Mendel, J.M., Nikias, C.L., 1998. Higher-order Spectral Analysis Toolbox for Use with Matlab. Http://www.mathworks.com

Taxt, T., 1997. Comparison of cepstrum-based methods for radial blind deconvolution of ultrasound images. *IEEE Trans. Ultrason. Ferroelect. Freq. Contr.*, **44**(3):666-674. [doi:10.1109/58.658327]

Wan, S., Raju, B.I., Srinivasan, M.A., 2003. Robust deconvolution of high-frequency ultrasound images using higher-order spectral analysis and wavelets. *IEEE Trans. Ultrason. Ferroelect. Freq. Contr.*, **50**(10):1286-1295. [doi:10.1109/TUFFC.2003.1244745]